\documentclass[10pt,twocolumn,letterpaper]{article}

\usepackage{cvpr}              %

\usepackage{_packages_}
\usepackage{_mathcmds_}
\usepackage{_envs_}
\usepackage{_cmds_}
\usepackage{_abbreviations_}
\usepackage{graphicx}
\usepackage{amsmath}
\usepackage{amssymb}
\usepackage{booktabs}

\usepackage{tabularx}
\usepackage{paralist}
\usepackage{listings}
\usepackage[newfloat,frozencache]{minted}

\definecolor{dkgreen}{rgb}{0,0.6,0}
\definecolor{gray}{rgb}{0.5,0.5,0.5}
\definecolor{mauve}{rgb}{0.58,0,0.82}
\makeatletter
\renewcommand\paragraph{\@startsection{paragraph}{4}{\z@}%
                                    {1.25ex \@plus1ex \@minus.2ex}%
                                    {-1em}%
                                    {\normalfont\normalsize\bfseries}}
\makeatother

\usepackage[pagebackref,breaklinks,colorlinks]{hyperref}
\hypersetup{
    colorlinks=true,
    linkcolor=blue,
    filecolor=blue,      
    urlcolor=blue,
    citecolor=blue
}

\usepackage[capitalize]{cleveref}
\crefname{section}{Sec.}{Secs.}
\Crefname{section}{Section}{Sections}
\Crefname{table}{Table}{Tables}
\crefname{table}{Tab.}{Tabs.}

\newcommand{\bla}[1]{#1}
\def\cvprPaperID{41} %
\def\confName{CVPR}
\def\confYear{2023}

\newcommand*{\myparagraph}[1]{\paragraph{#1}}
\newcommand{\cl}{Continual Learning\xspace}
\newcommand{\clshort}{CL\xspace}
\newcommand{\todo}[1]{\footnote{\textcolor{blue}{TODO: #1}}}

\newenvironment{code}{\captionsetup{type=listing}}{}
\SetupFloatingEnvironment{listing}{name=Code}
\DeclareFloatingEnvironment[
  fileext=loc,
  listname=List of codes,
  name=Code,
  placement=b,
]{codeb}
\DeclareFloatingEnvironment[
  fileext=loc,
  listname=List of codes,
  name=Code,
  placement=t,
]{codet}
\BeforeBeginEnvironment{minted}{\vspace*{-10pt}}

\begin{document}

\title{SequeL: A Continual Learning Library in PyTorch and JAX}

\author{Nikolaos Dimitriadis\\
EPFL\\
{\tt\small nikolaos.dimitriadis@epfl.ch}
\and
François Fleuret\\
University of Geneva\\
{\tt\small francois.fleuret@unige.ch}
\and
Pascal Frossard\\
EPFL\\
{\tt\small pascal.frossard@epfl.ch}
}
\maketitle

\begin{abstract}
Continual Learning is an important and challenging problem in machine learning, where models must adapt to a continuous stream of new data without forgetting previously acquired knowledge. While existing frameworks are built on PyTorch, the rising popularity of JAX might lead to divergent codebases, ultimately hindering reproducibility and progress. To address this problem, we introduce SequeL, a flexible and extensible library for Continual Learning that supports both PyTorch and JAX frameworks. SequeL provides a unified interface for a wide range of Continual Learning algorithms, including regularization-based approaches, replay-based approaches, and hybrid approaches. The library is designed towards modularity and simplicity, making the API suitable for both researchers and practitioners. We release SequeL\footnote{\url{https://github.com/nik-dim/sequel}} as an open-source library, enabling researchers and developers to easily experiment and extend the library for their own purposes.
\end{abstract}

\section{Introduction}
\label{sec:intro}
The field of Continual Learning (\clshort), also known as Lifelong Learning \cite{Parisi_Kemker_Part_Kanan_Wermter_2019}, Incremental Learning \cite{Rebuffi_Kolesnikov_Sperl_etal_2017}, or Sequential Learning, has seen fast growth in recent years. Continual Learning addresses the important setting of incrementally learning from a stream of data sources, disposing of the long-standing i.i.d. assumption of traditional machine learning. However, the pace of innovation has led to diverging settings in terms of datasets, assumptions, and requirements. As a consequence, several works have attempted to unify the Continual Learning paradigms \cite{Aljundi_Babiloni_Elhoseiny_etal_2018,Lopez-Paz_Ranzato_2017}.

The plethora of Continual Learning settings is accompanied by a variety of Deep Learning libraries, such as PyTorch, JAX, and TensorFlow, leading to further division. Each Deep Learning library has different advantages, and researchers opt for the one that better suits their needs and prior experience. Over time, the influx of new methods results in  disconnected repositories, stagnating progress due to limited reusability and lack of reproducibility.

\begin{codeb}
    \begin{minted}[
        frame=lines,
        fontsize=\footnotesize,
        escapeinside=||
    ]{python}
import sequel
import optax 
# PyTorch -> from torch.optim import optim

benchmark = sequel.benchmarks.SplitMNIST(
    num_tasks=5, batch_size=10)
    
model = sequel.backbones.|\colorbox{green}{jax}|.MLP(
    widths=[200], num_classes=10)
optimizer = optax.sgd(learning_rate=0.1)
# for PyTorch -> optimizer = optim.SGD(lr=0.1)

algo = sequel.algorithms.|\colorbox{green}{jax}|.EWC(
    model, benchmark, optimizer, 
    callbacks=[
        sequel.callbacks.|\colorbox{green}{Jax}|MetricCallback(), 
        TqdmCallback()], 
    loggers=[
        WandbLogger(...), CometLogger(...)],
    # additional arguments specific to EWC
    ewc_lambda=1,
    # optional arguments for all algorithms
    lr_decay=0.8)

algo.fit(epochs_per_task=1)
    \end{minted}
    \captionof{listing}{Example of an experiment in \sequel. By changing the highlighted \texttt{jax} to \texttt{pytorch}, and modifying the optimizer definition, the user can opt to run the experiment in either framework. }
    \label{code:example}
\end{codeb}

In this work, we propose \sequel, i.e., \underline{Seque}ntial \underline{L}earning, a Continual Learning framework written in both PyTorch and JAX. \sequel aims to unite the divergent codebases while allowing researchers to prototype fast without delving into engineering code, e.g., training loops and metric tracking. Users can develop in the framework of their choosing while accessing  the already implemented baselines. For example, consider the case where one researcher wants to implement a novel algorithm in JAX, but all the baselines are in PyTorch. Reimplementing everything from scratch is time-consuming and prone to mistakes. Instead, they can use our proposed framework to integrate their method and compare with baselines in an equal footing. 

Overall, \sequel offers a unified and flexible framework for Continual Learning research, which is easily extensible and  accessible in order to foster reproducibility. We believe \sequel  can help researchers to better compare methods and scale up to more complex Continual Learning settings.

\begin{figure*}[t]
    \centering
    \includegraphics[width=0.85\textwidth]{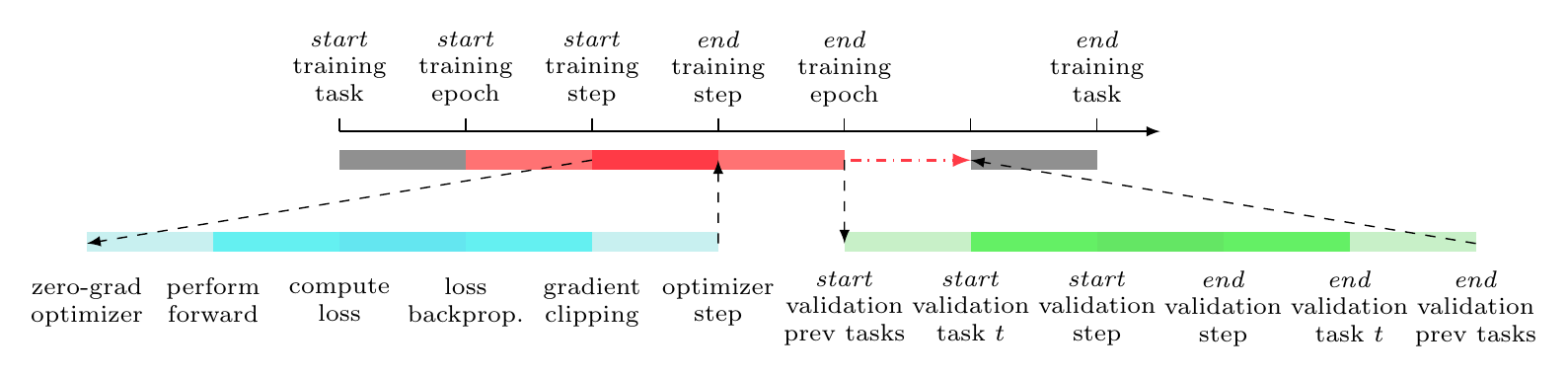}
    \caption{Control flow of the fitting process. Every point is surrounded by callback hooks. For instance, \texttt{training\_step()} is preceded by \texttt{on\_before\_training\_step()} and \texttt{on\_before\_training\_step\_callbacks()}  and proceeded by \texttt{on\_after\_training\_step()} and \texttt{on\_after\_training\_step\_callbacks()}.} 
    \label{fig:timeline}
\end{figure*}

\section{Framework}
\label{sec:framework}

The framework contains the following modules:
Benchmarks (\autoref{sec:benchmarks}),
Backbones (\autoref{sec:backbones}),
Callbacks (\autoref{sec:callbacks}),
Loggers (\autoref{sec:loggers}),
and
Algorithms (\autoref{sec:algos}). \autoref{code:example} shows the interplay between the different modules; a benchmark and a backbone along with loggers are fed into the algorithm instance that is endowed with additional custom functionalities via callbacks.
The algorithm serves as a trainer module and also houses the conceptual details of \clshort methodologies, such as Averaged-Gradient Episodic Memory \cite{Chaudhry_Ranzato_Rohrbach_etal_2019}. 
\sequel provides flexibility by allowing users to develop their algorithm in either \jax or \pytorch. The framework has been geared towards ease-of-use and division of engineering and algorithmic details.

\subsection{Benchmarks}
\label{sec:benchmarks}

The \textit{Benchmarks} module provides several widely-used \cl benchmarks, both in the New Instance (NI) and New Class (NC) scenarios. The currently supported benchmarks are Split/Permuted/Rotated MNIST, Split CIFAR10/100 and Split TinyImageNet.
The \textit{Benchmarks} module is implemented in \pytorch, since it is better suited for dynamically generating and handling data streams. During training, the input and targets are transformed to the appropriate format to ensure compatibility with both \pytorch and \jax.  
All supported benchmarks are based on \texttt{BaseBenchmark} class that handles most use cases such as loading training and validation streams for task \( t \), for all tasks up to \( t \), i.e., \( \{1,2,\dots ,t \} \). Similar functionalities are provided for memory capabilities, such as loading memory streams for one or more tasks, augmenting the current task dataset with the memories of all previous tasks etc. The benchmark module also handles dataloader construction, and the input to the algorithm is a tuple \( (x, y, t) \) of inputs \( x \), targets \( y \) and task IDs \( t \).

\myparagraph{Implementing new benchmarks} Let \( \numtasks \) be the number of tasks. Each Benchmark must implement the method \texttt{prepare\_datasets} that returns two dictionaries of \( \numtasks \) key-value pairs that contain the dataset for the corresponding task \( t\in[T] \), for training and validation. For New Class scenarios, such as SplitMNIST or SplitCIFAR100, the method creates a disjoint datasets and for New Instance scenarios, such as PermutedMNIST or RotatedMNIST, each dataset is coupled with a specific \texttt{torchvision} transformation, regarding the fixed permutation or rotation.

\subsection{Models/Backbones}  
\label{sec:backbones}
The \textit{Models} module contains neural networks widely used in the literature, such as MultiLayer Perceptrons and Convolutional Neural Networks, Residual Networks \cite{He_Zhang_Ren_etal_2016}. For both \jax and \pytorch, a \texttt{BaseBackbone} class is defined that inherits from \texttt{flax.nn.Module} and \texttt{torch.nn.Module}, respectively and endows the model with the functionality of selecting the output head for NC benchmarks. The user can easily extend the module with models stemming from the literature or that are custom-made by changing the base class to \texttt{BaseBackbone}. A utility model is also provided that receives as input a \texttt{torch/flax.nn.Module} and wraps it with the appropriate \texttt{BaseBackbone} to facilitate importability.

\subsection{Callbacks} 
\label{sec:callbacks}

A callback provides hooks for any point in training and validation, similar to Pytorch Lightning \cite{Falcon_PyTorch_Lightning_2019}. It offers the ability to extend or probe the algorithm and/or model during fitting. Metric callbacks have been implemented for both JAX and PyTorch and handle the monitoring of metrics, e.g., accuracy and forgetting, and the ad hoc tracking via the \textit{Loggers} module. 
Utility callbacks can be implemented, such as the \texttt{TqdmCallback} that provides additional information during training via a progress bar.

\subsection{Loggers}     
\label{sec:loggers}

Experiment tracking has become an indispensable part of the \mlshort pipeline. Hence, \sequel includes five different loggers: \texttt{LocalLogger}, \texttt{ConsoleLogger}, \texttt{WandbLogger}\footnote{Wandb refers to Weights \& Biases \cite{wandb}}, \texttt{TensorBoardLogger} and \texttt{CometLogger}, allowing users to track their runs with the preferred service. Specifically, \texttt{LocalLogger} saves the evolution and final metrics in a local file, while \texttt{ConsoleLogger} prints information as a table in the console. \texttt{WandbLogger}, \texttt{TensorBoardLogger} and \texttt{CometLogger} use the APIs of the homonym services, allowing the tracking of images, tables etc. and the integration with powerful visualization tools and dashboards.

\subsection{Algorithms}
\label{sec:algos}

The \textit{Algorithms} module controls the program flow and incorporates all the aforementioned modules. By calling the \texttt{fit} method, training with validation occurs for the selected \bla{backbone} for the given \bla{benchmark}, tracking metrics via the corresponding \bla{callback} and logging them to the desired service via a \bla{logger}. The parent class \bla{BaseAlgorithm} handles the engineering code, while the algorithmic parts are implemented by the children classes. This design choice is motivated by the desire to have access to all internal variables, such as the input \( x \) and task ID \( t \) of the current batch, without using a separate training module. As a result, the engineering logic is kept separate from research code via inheritance. \autoref{fig:timeline} shows a simplified version of the program flow. Each event is encircled by the homonym callbacks.

The \texttt{BaseAlgorithm} class is framework agnostic and primarily sets the control flow of the program, such as training for one task and then validating current and preceding tasks. The peculiarities and design constraints imposed by the \pytorch and \jax philosophy are handled by the corresponding base classes, \texttt{PyTorchBaseAlgorithm}  and \texttt{JaxBaseAlgorithm}. For instance, for \pytorch the current batch is moved to the appropriate CUDA device, while for JAX it is converted to the NumPy format.

The \texttt{BaseAlgorithm} offers basic functionality and uses callbacks for specific and custom functionalities. It also inherits from \texttt{BaseCallback} and provides the same hooks outlined in \autoref{sec:callbacks}. Overall, each event \( \texttt{E}  \) in training, e.g., \texttt{training\_epoch()} is surrounded by four hooks in the following sequence: \texttt{on\_before\_E}, \texttt{on\_before\_E\_callbacks}, \texttt{E}, \texttt{on\_after\_E} and \texttt{on\_after\_E\_callbacks}. 
Hence, the user can choose to implement an algorithm via specific callbacks or in child classes so that the research code is concentrated in a single file. For improved readability, the currently supported methods opt for the latter. Calculation of metrics, utilities for printing to console etc. are reserved for callbacks and the corresponding hooks.

The framework includes implementations for 
Naive Finetuning, 
Elastic Weight Consolidation (EWC) \cite{Kirkpatrick_Pascanu_Rabinowitz_etal_2017}, 
Synaptic Intelligence (SI) \cite{Zenke_Poole_Ganguli_2017}, 
Memory Aware Synapses (MAS) \cite{Aljundi_Babiloni_Elhoseiny_etal_2018}, 
Averaged-Gradient Episodic Memory (A-GEM) \cite{Chaudhry_Ranzato_Rohrbach_etal_2019}, 
Less-Forgetting Learning (LFL) \cite{Jung_Ju_Jung_Kim_2018},
Experience Replay (ER) \cite{Chaudhry_Rohrbach_Elhoseiny_Ajanthan_Dokania_Torr_Ranzato_2019},
Dark Experience Replay (DER and DER++) \cite{Buzzega_Boschini_Porrello_Abati_Calderara_2020},
Stable SGD \cite{Mirzadeh_Farajtabar_Pascanu_Ghasemzadeh_2020},
Kernel Continual Learning (KCL) \cite{Derakhshani_Zhen_Shao_etal_2021},
Look Ahead Model Agnostic Meta Learning (LaMAML) \cite{Gupta_Akin_2020},
and
Mode Connectivity Stochastic Gradient Descent (MC-SGD) \cite{Mirzadeh_Farajtabar_Gorur_etal_2021}.

\textbf{Implementing new algorithms}
\sequel supports regularization and replay algorithms, via out-of-the-box components. Parent classes are implemented for the specific realizations of regularization-based algorithms such as Elastic Weight Consolidation. 
For replay methods, the \texttt{MemoryMechanism} class and the corresponding callback handle saving samples in the memory and their selection process. 
For regularization algorithms, the overall loss for sample \( (\xx, y, t) \) of a classification problem is \( 
    \calL (\xx, y) = \calL_{CE} (f_{\bm{\theta}}(\xx), y) + \lambda \sum_{i} \Omega_i (\theta_i - \theta_{i, \textrm{old}})^2  
     \)
where \( f \) is a neural network parameterized by \( \bm{\theta} \), \( \bm{\theta}_{\textrm{old}} \) are the parameters at the end of training of the previous task, \( \Omega_i \) refers to the  importance of parameter \( i \) and \( \lambda \) is the regularization coefficient. To add a new regularization method, the user needs only to implement the \texttt{calculate\_parameter\_importance} method to calculate \(  \Omega_i  \), while the storing of the old parameters and the calculation of the regularization loss is handled by the parent class. In case of algorithms such as Synaptic Intelligence \cite{Zenke_Poole_Ganguli_2017} that keep an online internal parameter \( \omega_i \) that is later used to compute \( \Omega_i \), the method \texttt{on\_after\_training\_step} houses the corresponding algorithmic details.

\myparagraph{Reproducibility}
To encourage transparency, \sequel uses Hydra \cite{Yadan2019Hydra} configuration files to formalize experiments. While an experiment can be constructed as in \autoref{code:example}, an alternative lies in defining a configuration file, as in \autoref{code:hydra config}. Instead of obfuscating hyperparameters and impeding reproducible results, an experiment defined with Hydra can be easily shared and reported. This feature is enabled by a series of routers that select the correct benchmark or model, and the implementation of the \texttt{from\_config} method for all related module classes. 
\sequel includes such configuration files reproducing \cl baselines reported in various papers. Example configuration files along with reproducibility runs tracked via Weights\&Biases are provided; the experiments focus on \texttt{RotatedMNIST} and include  classic algorithms, such as EWC \cite{Kirkpatrick_Pascanu_Rabinowitz_etal_2017} and Naive SGD, as well as more involved baselines in MCSGD \cite{Mirzadeh_Farajtabar_Gorur_etal_2021} and LaMAML \cite{Gupta_Akin_2020}. The list will be expanded to ensure correctness. See \autoref{appendix:Reproducibility} for more details.

\myparagraph{Hyperparameter Tuning}
Another benefit of the Hydra-based \cite{Yadan2019Hydra} setup  is its out-of-the-box hyperparameter tuning capabilities, allowing for the quick setup of ablation studies. Specifically, the user picks as a basis the aforementioned config file and defines the settings of a grid search, such as \texttt{batch\_size} \( \in \{10, 20, 30\} \) and \texttt{lr} \( \in \{0.01, 0.1\} \).

\begin{codeb}
    \begin{minted}[
        frame=lines,
        fontsize=\footnotesize,
    ]{yaml}
version: 0.0.1 # of SequeL framework 
mode: pytorch # or jax
algo:
    name: ewc
    ewc_lambda: 1.0
benchmark:
    name: rotatedmnist
    batch_size: 10
    num_tasks: 20
    per_task_rotation: 9
backbone:
    type: mlp
    n_hidden_layers: 2 
    width: 256
    num_classes: 10
    dropout: 0.2
optimizer:
    type: sgd
    lr: 0.01
    lr_decay: 0.8
training:
    epochs_per_task: 1
wandb: # enables Weights and Biases tracking 
    entity: ENTITY
    project: PROJECT
    \end{minted}
    \captionof{listing}{Example of a configuration file. Experiments in \sequel can be fully expressed and started easily by defining the settings in a yaml file. The configuration file includes the version of the framework as well, in this case \texttt{0.0.1}.}
    \label{code:hydra config}
\end{codeb}

\section{Related Work}  

The progress of the Machine Learning community can be attributed to large extent to the development of Deep Learning libraries, such as \pytorch \cite{pytorch}, \tensorflow \cite{tensorflow2015-whitepaper} and \jax \cite{jax2018github}, which abstract low-level engineering code and provide a high-level API to the user. Thus, researchers and practitioners can reliably develop new methodologies by focusing on the algorithmic inner workings. 

The progress of the field in conjunction with the fact that majority of the \mlshort pipelines are similar has pushed for the creation of frameworks that provide further abstractions. Pytorch Lightning \cite{Falcon_PyTorch_Lightning_2019} and fastai\cite{fastai} are general \mlshort libraries that extend flexibility via a wide range of callbacks and loggers, while minimizing the engineering overhead. 

The progressive increase in abstraction has led to the development of libraries specialized towards specific \mlshort subfields and that use the aforementioned software as building blocks. For instance,  the HuggingFace\cite{Wolf_Transformers_State-of-the-Art_Natural_2020} library for Transformers \cite{Vaswani_Shazeer_Parmar_etal_2017}  includes pretrained models for Natural Language Processing and, more recently, Computer Vision \cite{Dosovitskiy_Beyer_Kolesnikov_Weissenborn_Zhai_Unterthiner_Dehghani_Minderer_Heigold_Gelly_et_al_2021}. PyTorch Geometric \cite{Fey_Fast_Graph_Representation_2019} offers a comprehensive suite of tools geared towards Graph Neural Networks (GNNs). Deep Graph Library \cite{wang2019dgl} also focuses on deep learning for graphs and is also framework agnostic, i.e., it supports PyTorch, Apache MXNet and TensorFlow. \sequel shares this trait and offers the user the flexibility of two ecosystems in PyTorch and JAX. 
Multiple packages exist in the Reinforcement Learning literature, e.g., OpenAI Gym\cite{openai-gym} and OpenAI baselines \cite{baselines}.
MMSegmentation\cite{MMSegmentation_Contributors_OpenMMLab_Semantic_Segmentation_2020} and Segmentation-Models-PyTorch \cite{smp} are semantic segmentation toolboxes. \sequel's Hydra integration  
shares the design philosophy of the configuration files used in the former.

An important aspect of the \mlshort toolbox focuses on experiment tracking and monitoring and is becoming more important given the increasing complexity of models and methods, the need of rigorous ablation studies and hyperparameter tunings. Several frameworks address these imperatives, such as  
MLFlow \cite{Chen_Chow_Davidson_DCunha_Ghodsi_Hong_Konwinski_Mewald_Murching_Nykodym_et}, Weights and Biases \cite{wandb}, Comet and TensorBoard \cite{tensorflow2015-whitepaper}. \sequel incorporates the logging capabilities of such libraries and allows users to track and visualize their experiments with the service of their choosing.

The overarching effort to create easy-to-use and reliable tools has also been observed in the \cl realm. Several libraries have been proposed, such as Avalanche\cite{Larsson_Maire_Shakhnarovich_2017}, CL-Gym \cite{Mirzadeh_2021_CVPR} and  Sequoia \cite{Normandin_Golemo_Ostapenko_etal_}. Avalanche and CL-Gym share a similar design in terms of module structure and focus on the supervised setting. The algorithm selection in CL-Gym is limited and, while Avalanche offers a wide range of algorithms, the focus lies on more classical algorithms. For instance, Kernel Continual Learning \cite{Derakhshani_Zhen_Shao_etal_2021} and Dark Experience Replay \cite{Buzzega_Boschini_Porrello_Abati_Calderara_2020} are not implemented in either framework.
 AvalancheRL\cite{avalanche_rl} extends Avalanche with functionalities for Reinforcement Learning. Sequoia \cite{Normandin_Golemo_Ostapenko_etal_} focuses on the Reinforcement Learning perspective of \cl and uses components of OpenAI Gym \cite{openai-gym}, Avalanche \cite{Lomonaco_Avalanche_an_End-to-End_2021} and Continuum \cite{douillardlesort2021continuum}. Compared to the aforementioned libraries, \sequel supports both \pytorch and \jax, simplifying the comparison and importability of novel methods and implementations of existing approaches irrespective of framework.

\section{Conclusion}
\label{sec:conclusion}

In conclusion, we have presented \sequel, a novel \cl framework written in both PyTorch and JAX, aimed at unifying divergent codebases and facilitating reproducible research in the field of \cl. Our library provides a convenient and flexible platform for researchers to prototype and test their novel algorithms, as well as compare them to existing state-of-the-art methods. We believe that our library will contribute to the growth of \cl research and provide a valuable resource for the community.

{\small
\bibliographystyle{ieee_fullname}
\bibliography{references}
}

\clearpage
\appendix

\section{Reproducibility}
\label{appendix:Reproducibility}

\sequel offers  configuration files for reproducibility purposes and to ensure the correctness of the framework. 
Two examples based on experimental results reported for \bla{RotatedMNIST} \cite{Mirzadeh_Farajtabar_Gorur_etal_2021,Gupta_Akin_2020} are shown in \autoref{config:mcsgd} and \autoref{config:lamaml}. Instructions on how to run the experiments along with more examples are provided in the repository.

\begin{code}
    \begin{minted}[
        frame=lines,
        fontsize=\footnotesize,
    ]{yaml}
version: 0.0.1
expected:
  avg_acc: 82.3 # as reported in the paper
benchmark:
  name: rotatedmnist
  num_tasks: 20
  per_task_rotation: 9
  batch_size: 64
  eval_batch_size: 1024
source: mcsgd paper
algo:
  name: mcsgd
  per_task_memory_samples: 100
  lmc_policy: offline
  lmc_interpolation: linear
  lmc_lr: 0.05
  lmc_momentum: 0.8
  lmc_batch_size: 64
  lmc_init_position: 0.1
  lmc_line_samples: 10
  lmc_epochs: 1
  lr_decay: 0.8
backbone:
  type: mlp
  n_hidden_layers: 2
  width: 256
  num_classes: 10
  dropout: 0.2
optimizer:
  type: sgd
  lr: 0.1
  momentum: 0.8
training:
  epochs_per_task: 1
\end{minted}
\captionof{listing}{Reproducibility Experiment for MCSGD.}
\label{config:mcsgd}
\end{code}

\newpage

\begin{code}
    \begin{minted}[
        frame=lines,
        fontsize=\footnotesize,
    ]{yaml}
version: 0.0.1
expected:
  avg_acc: 77.42 # as reported in the paper
source: original paper
algo:
  name: lamaml
  glances: 5
  n_inner_updates: 5
  second_order: false
  grad_clip_norm: 2.0
  learn_lr: true
  lr_alpha: 0.3
  sync_update: false
  mem_size: 200
backbone:
  type: mlp
  n_hidden_layers: 2
  width: 100
optimizer:
  type: sgd
  lr: 0.1
benchmark:
  name: rotatedmnist
  per_task_rotation: 9
  num_tasks: 20
  batch_size: 10
  eval_batch_size: 10000
  subset: 1000
training:
  epoch_per_task: 1
\end{minted}
\captionof{listing}{Reproducibility Experiment for LaMAML.}
\label{config:lamaml}
\end{code}

\end{document}